\documentclass[10pt,twocolumn,letterpaper]{article}

\usepackage[pagenumbers]{cvpr} 

\usepackage{graphicx}       
\usepackage{amsmath}        
\usepackage{amssymb}        
\usepackage{amsfonts}       
\usepackage{booktabs}       
\usepackage{multirow}       

\usepackage{verbatim}
\usepackage{algorithm}
\usepackage[noend]{algpseudocode}

\usepackage{subcaption}
\usepackage{float}
\usepackage{caption}
\usepackage{pifont}

\usepackage{xcolor}  
\usepackage[table]{xcolor}
\definecolor{mygray}{RGB}{220,220,220}  
\usepackage{fontawesome5}

\definecolor{cvprblue}{rgb}{0.21,0.49,0.74}
\usepackage[pagebackref,breaklinks,colorlinks,allcolors=cvprblue]{hyperref}

\def\paperID{25} 
\def\confName{CVPR}
\def\confYear{2026}

\title{Can We Go Beyond Visual Features? Neural Tissue Relation Modeling for Relational Graph Analysis in Non-Melanoma Skin Histology}
\author{
Shravan Venkatraman\textsuperscript{1,2} \quad
Muthu Subash Kavitha\textsuperscript{2} \quad
Joe Dhanith P R\textsuperscript{3} \\
Manikandarajan Venmathimaran\textsuperscript{4} \quad
Jia Wu\textsuperscript{5} \\
\textsuperscript{1}Mohamed bin Zayed University of AI, Abu Dhabi, UAE \\
\textsuperscript{2}School of Information and Data Sciences, Nagasaki University, Nagasaki, Japan \\
\textsuperscript{3}Vellore Institute of Technology, Chennai, India \quad
\textsuperscript{4}Loughborough University, United Kingdom \\
\textsuperscript{5}MD Anderson Cancer Center, The University of Texas, Houston, USA \\
{\small \faEnvelope \hspace{5pt} \texttt{shravan.venkatraman@mbzuai.ac.ae, kavitha@nagasaki-u.ac.jp}}
}
\begin{document}
\maketitle


\begin{abstract}
Histopathology image segmentation is essential for delineating tissue structures in skin cancer diagnostics, but modeling spatial context and inter-tissue relationships remains a challenge, especially in regions with overlapping or morphologically similar tissues. Current convolutional neural network (CNN)- and Transformer-based approaches operate primarily on visual texture, often treating tissues as independent regions and failing to encode biological context. To this end, we introduce Neural Tissue Relation Modeling (NTRM), a novel segmentation framework that augments CNNs with a tissue-level graph neural network to model spatial and functional relationships across tissue types. NTRM constructs a graph over predicted regions, propagates contextual information via message passing, and refines segmentation through spatial projection. Unlike prior methods, NTRM explicitly encodes inter-tissue dependencies, enabling structurally coherent predictions in boundary-dense zones. On the benchmark Histopathology Non-Melanoma Skin Cancer Segmentation Dataset, NTRM outperforms state-of-the-art methods, achieving a robust Dice similarity coefficient that is 4.9\% to 31.25\% higher than the best-performing models among the evaluated approaches. Our experiments indicate that relational modeling offers a principled path toward more context-aware and interpretable histological segmentation, compared to local receptive-field architectures that lack tissue-level structural awareness. Our code is available \href{https://shravan-18.github.io/NTRM/}{\textcolor{magenta}{here}}.
\end{abstract}    

\section{Introduction}
\label{sec:introduction}

Non-melanoma skin cancers, including basal cell carcinoma and squamous cell carcinoma, remain the most common malignancies worldwide, with recent estimates indicating over 1.2 million new cases diagnosed globally in 2022 and a continued upward trend through 2025~\cite{frontiers2025,wang2025,yan2023cvpr,takt}. Accurate histopathological diagnosis requires precise identification and delineation of distinct tissue types within complex microenvironments~\cite{sirinukunwattana2016,miccaihistopath2}. Current diagnostic workflows depend heavily on pathologist expertise to interpret spatial relationships between tissue components, yet these critical contextual dependencies remain largely unaddressed by existing computational approaches~\cite{coppola2020cvprw,miccaihistopath1}.

\begin{figure*}[t]
    \centering
    \includegraphics[width=1\textwidth]{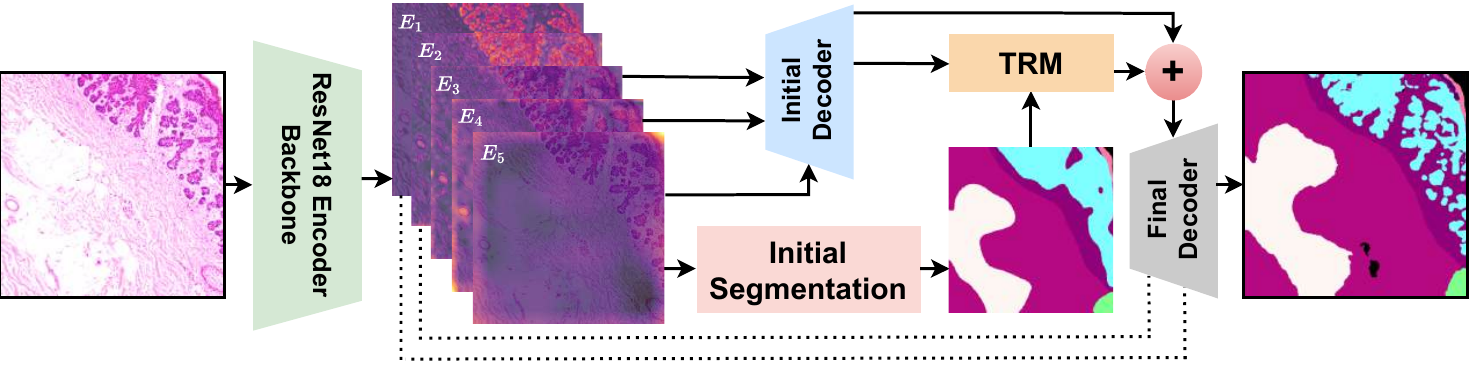}
    \caption{NTRM framework pipeline showing CNN-based encoding, initial segmentation, TRM module, and final decoding for relationally-informed histological segmentation.}
    \label{fig:pipeline}
\end{figure*}

Deep learning methods have become foundational in histopathology image analysis, enabling accurate tissue identification \cite{litjens2016deep} and segmentation \cite{ronneberger2015u,graham2019hover,thomasBase,caNet2021}. Attention mechanisms enhance these models by focusing on clinically salient regions \cite{ilse2018attention}. However, despite their strength in capturing texture and morphology, CNN- and transformer-based approaches treat pixels or patches independently, neglecting the inter-tissue spatial dependencies and critical contextual information that emerge from tissue-tissue interactions \cite{hybrid2025}.

While recent methods incorporate spatial context via multi-scale architectures \cite{chen2017deeplab,structureAware} and cell-level graph representations \cite{graphRep}, the modeling of inter-tissue relationships remains largely overlooked in histopathology segmentation \cite{miccaihistopath3,miccaihistopath4,miccaihistopath5}. Existing graph neural networks (GNNs) focus on cell-cell interactions rather than tissue-level dependencies \cite{cgcom}, and segmentation methods typically treat tissues in isolation without considering their biological co-occurrence patterns or spatial dependencies \cite{imran2024transformer,dynamiccvpr}. This limitation becomes particularly problematic when distinguishing between morphologically similar tissues that differ primarily in their biological context and relationships to neighboring structures \cite{relationalGCN}.

We address this gap by introducing Neural Tissue Relational Modeling (NTRM), a novel framework that explicitly models the biological relationships between tissue types through a GNN integrated with traditional CNN feature extraction. Our approach constructs a tissue-level graph where nodes represent different tissue types and edges encode their spatial and functional relationships, learning tissue-specific embeddings that capture both visual characteristics and biological context. We do this by combining an initial draft segmentation with a tissue relation module (TRM) that refines predictions by incorporating learned tissue dependencies, as illustrated in Fig.~\ref{fig:pipeline}.

To summarize, we make the following contributions:
\begin{itemize}
    \item A novel framework for modeling inter-tissue biological relationships through spatial-functional graphs in histology images.
    \item A TRM that learns tissue-specific embeddings and integrates relational knowledge with CNN features to refine segmentation predictions.
    \item A region-based graph construction strategy that handles irregular tissue shapes through masked pooling and incorporates global tissue knowledge embeddings for missing tissue types.
\end{itemize}

\section{Related Work}
\label{sec:relatedWorks}

\paragraph{Deep Learning for Histopathology Segmentation.}

Deep learning has become the dominant paradigm for histopathology image segmentation, enabling automated analysis of complex tissue structures and cellular patterns that are critical for diagnosis and prognosis \cite{s1p5}. Early works primarily relied on convolutional architectures to learn pixel-level representations from histology images. For instance, Naylor et al. formulate nuclei segmentation as a distance-map regression task using fully convolutional networks to better separate touching nuclei \cite{s1p6}. Subsequent approaches have focused on improving feature representations and segmentation robustness. Deep WSI-Stroma employs multiscale CNN-based learning to segment epithelial and mesenchymal regions in breast cancer slides \cite{s1p7}, while StarDist-based nuclei detection frameworks adapt pretrained models to domain-specific staining conditions to achieve reliable cellular segmentation in clinical workflows \cite{s1p4}. To reduce the dependence on dense annotations, weakly supervised methods have also been explored; Zhang et al. introduce a contrast-based variational framework that leverages sparse point annotations to generate complementary supervision for segmentation networks \cite{s1p8}. Similarly, diffusion-based approaches have recently emerged as powerful representation learners. GenSelfDiff-HIS employs generative diffusion models within a self-supervised learning framework to address the scarcity of labeled histopathology data \cite{s1p9}, while PathSegDiff utilizes pathology-specific latent diffusion representations as pretrained feature extractors to improve segmentation accuracy in tumor and tissue classification tasks \cite{s1p1}. Hybrid architectures have also been proposed to combine complementary modeling capabilities; ACS-SegNet integrates CNN and vision transformer encoders with attention-driven feature fusion for improved tissue segmentation \cite{s1p2}, and ClinSegNet introduces a recall-oriented optimization strategy with uncertainty-driven refinement to better detect small lesions and indistinct boundaries \cite{s1p3}. Despite these advances in representation learning, supervision strategies, and architectural design, most existing methods still treat segmentation primarily as a pixel-level prediction problem based on visual features alone, without explicitly modeling the structured relationships between different tissue regions that often govern histological organization.

\begin{figure*}[t]
  \centering
  \includegraphics[width=\textwidth]{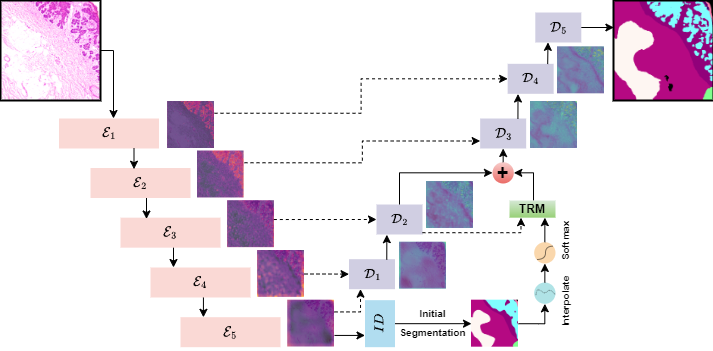}
  \caption{\textbf{NTRM} architecture. A ResNet18 backbone extracts hierarchical encoder features $\{\mathcal{E}_1,\dots,\mathcal{E}_5\}$, which are decoded into an initial segmentation map. The TRM module receives this map and early decoded features $\mathcal{D}_2$, and refines them via graphical modeling of tissue-type relationships. The final prediction is produced after fusing the refined features with $\mathcal{D}_2$ via deeper decoder layers.}
  \label{fig:architecture}
\end{figure*}

\paragraph{Attention and Transformer-Based Context Modeling.}

Transformer architectures have emerged as a powerful paradigm for modeling contextual relationships through self-attention mechanisms, enabling dynamic interactions between tokens or spatial regions. Recent research has focused on improving the scalability, efficiency, and interpretability of attention-based models. For instance, SLAY introduces a geometry-aware linearized attention mechanism based on the Yat-kernel, enabling near-softmax performance while achieving linear-time complexity for long-context modeling \cite{s2p1}. Hardware-aware optimizations have also been explored; specialized FlashAttention accelerators fuse exponential and multiplication operations to significantly reduce power and area costs in attention computation \cite{s2p2}. From a theoretical perspective, recent studies analyze the representational properties of transformers, demonstrating that softmax-based attention can achieve universal consistency for functional regression tasks under appropriate conditions \cite{s2p3}. Other work addresses the challenge of long-context reasoning. Infini-attention augments the transformer with compressive memory and hybrid local–global attention, enabling efficient modeling of extremely long sequences with bounded memory requirements \cite{s2p4}.

Beyond sequence modeling, attention mechanisms have also been widely adopted in computer vision. Vision transformers replace convolution with attention to capture global context, and architectures such as DeBiFormer introduce deformable bi-level routing attention to improve the selection of relevant key–value pairs during inference \cite{s2p5}. Recent theoretical studies further investigate how transformers encode latent causal structures through attention during gradient-based training \cite{s2p6}. Complementary approaches aim to improve the quality of contextual reasoning by suppressing irrelevant information: Selective Attention reduces unnecessary context interactions to improve efficiency and downstream performance \cite{s2p7}, while Selective Self-Attention introduces temperature-controlled sparsity to mitigate attention dilution and improve relevance modeling \cite{s2p8}. Despite these advances in contextual representation, attention-based models primarily operate over token- or patch-level interactions and do not explicitly encode higher-level semantic relationships between structured entities such as tissue regions in histopathology images.

\paragraph{Graph-Based Learning in Computational Pathology.}

Graph-based learning has emerged as an effective paradigm in computational pathology for modeling the structural organization and spatial dependencies that are often lost in patch-wise or bag-based representations. Recent work has extended graph reasoning across multiple levels of pathology analysis. At the whole-slide level, MMSF incorporates a graph feature extraction module to encode tissue topology at the patch level and fuse it with clinical information for classification and survival analysis \cite{s3p1}, while Geo-MIL formulates weakly supervised gastric cancer segmentation as graph-based multi-instance learning, explicitly modeling spatial relationships between tissue patches to bridge slide-level labels and pixel-level predictions \cite{s3p2}. Interpretable and structure-aware WSI modeling has also gained attention: biologically informed tissue-region graphs with graph attention have been used to achieve competitive diagnostic performance while preserving explainability \cite{s3p3}, and deformable-attention GNNs further enhance contextual understanding by constructing dynamic weighted graphs that incorporate learnable spatial offsets over tissue patches \cite{s3p4}. Other studies strengthen graph representations through improved features, multimodal fusion, or adaptive connectivity. Foundation-model features have been integrated into lightweight GNNs for histology classification, demonstrating the value of stronger node embeddings and complementary fusion with fine-tuned models \cite{s3p5}; SIGMMA extends graph learning to multimodal pathology by representing cell interactions as hierarchical graphs for joint histology--transcriptomics alignment \cite{s3p6}; and dynamic graph methods such as WiKG replace fixed spatial topologies with knowledge-aware neighbor construction and edge-aware attention for WSI classification \cite{s3p7}. 

Beyond slide-level modeling, graph reasoning has also been applied at the cellular and multi-graph levels. Cell Graph Transformer treats nuclei graphs as tokenized node-edge structures to enable learnable adjacency and global information exchange \cite{s3p8}, SCUBa-Net jointly models pathology images through spatially constrained and unconstrained graphs with inter- and intra-graph interaction blocks \cite{s3p9}, and SlideGCD incorporates inter-slide correlations through slide-based graph collaborative training to improve multiple WSI analysis tasks \cite{s3p10}. Despite these advances, most existing graph-based pathology methods operate at the level of cells, patches, or entire slides for classification, retrieval, or multimodal analysis, and thus do not directly address the problem of tissue-level relational reasoning within dense segmentation, where explicit modeling of dependencies among neighboring tissue regions is crucial for structurally coherent predictions.

\begin{figure*}[t]
  \centering
  \includegraphics[width=\textwidth]{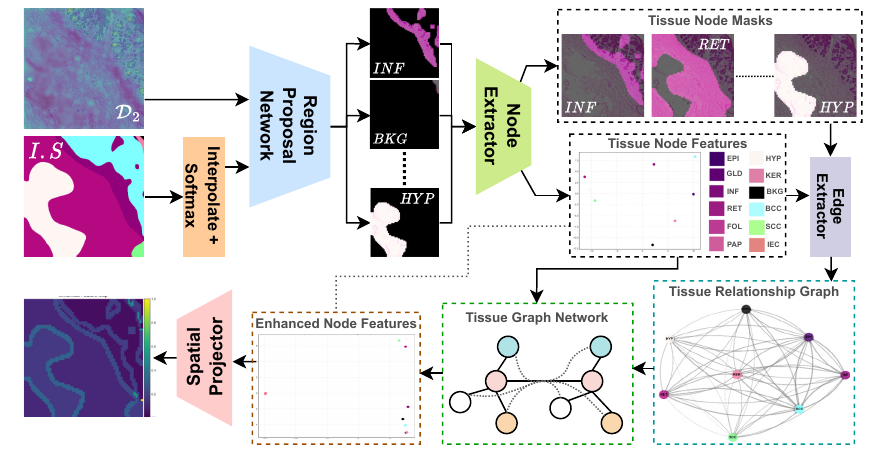}
  \caption{Pipeline of the TRM module. Initial softmax predictions and early CNN features are used to define tissue-specific regions. Node features are extracted via masked pooling, and edges are created between spatially adjacent regions. A GNN performs message passing over this tissue graph, and refined node embeddings are projected back to the spatial domain. Tissue visuals shown include INF (Inflammation), BKG (Background), RET (Reticular Dermis), and HYP (Hypodermis).}
  \label{fig:trm}
\end{figure*}

\section{Proposed Method: \textit{NTRM}}
\label{sec:proposed}

As shown in Fig.~\ref{fig:architecture}, our method consists of a ResNet-based encoder-decoder backbone \cite{resnet,ronneberger2015u}, an initial segmentation head, and a TRM that refines segmentation based on inferred spatial and functional tissue relationships. Let $x \in \mathbb{R}^{3\times H \times W}$ denote an input histology image. The encoder extracts features $\{\mathcal{E}_1, \mathcal{E}_2, \dots, \mathcal{E}_5\}$, where $\mathcal{E}_i \in \mathbb{R}^{C_i \times H_i \times W_i}$ represent hierarchical visual embeddings at different resolutions. Decoder blocks then aggregate these to generate intermediate features $\mathcal{D}_1$ and $\mathcal{D}_2$, where $\mathcal{D}_2 \in \mathbb{R}^{C \times H_2 \times W_2}$ provides the input to both the initial segmentation head and the TRM.

The initial decoder $\phi$ generates a coarse segmentation map $\hat{y}_{\text{init}} = \phi(\mathcal{E}_5) \in \mathbb{R}^{K \times H_0 \times W_0}$, where $K$ denotes the number of tissue classes. This map is bilinearly upsampled to match $H_2 \times W_2$ and passed through a softmax to produce class-wise probabilities $p \in \mathbb{R}^{K \times H_2 \times W_2}$, which serve as input to the TRM.

\subsection{Tissue Relation Module} 

The TRM constructs a tissue-level graph $\mathcal{G} = (\mathcal{V}, \mathcal{E})$ from the coarse segmentation map, where each node represents a predicted tissue class and edges indicate contextual or spatial proximity. Softmax-normalized class probabilities are thresholded to generate binary masks for each tissue, which define the spatial extent of each node. Intermediate CNN features $\mathcal{D}_2$ are then masked and globally pooled to produce class-specific node embeddings. Edges are constructed by examining spatial adjacency between tissue masks, allowing the graph to capture biologically relevant neighborhood relationships. This explicit graph representation enables TRM to reason over tissue co-occurrence and context - modeling structured interactions that convolutional layers alone cannot express. A graph neural network propagates messages over $\mathcal{G}$, refining node embeddings before projecting them back into the spatial domain.

\begin{algorithm}[t]
\caption{Tissue Graph Construction}
\label{alg:graph_construction}
\begin{algorithmic}[1]
\Require Initial segmentation probabilities $p \in \mathbb{R}^{K \times H \times W}$, features $\mathcal{D}_2 \in \mathbb{R}^{C \times H \times W}$, threshold $\tau$
\Ensure Tissue graph $\mathcal{G} = (\mathcal{V}, \mathcal{E})$, node features $H \in \mathbb{R}^{K \times d}$, edge features $E$
\State $\mathcal{V} \gets \{1, 2, \ldots, K\}$ \Comment{Initialize vertex set}
\State $\mathcal{E} \gets \emptyset$ \Comment{Initialize edge set}
\State $F \gets \psi(\mathcal{D}_2)$ \Comment{Apply convolutional projection}
\For{$c = 1$ to $K$}
    \State $M_c \gets \mathbb{I}[\text{maxpool}(p_c) > \tau]$ \Comment{Generate binary mask}
    \State $F_c \gets F \odot M_c$ \Comment{Mask features}
    \State $h_c \gets \frac{\sum_{i,j} F_c(i,j)}{\sum_{i,j} M_c(i,j) + \varepsilon}$ \Comment{Compute node embedding}
    \If{$\sum_{i,j} M_c(i,j) = 0$}
        \State $h_c \gets g_c$ \Comment{Use global embedding if class is absent}
    \EndIf
\EndFor
\For{$i = 1$ to $K$}
    \For{$j = 1$ to $K$, $j \neq i$}
        \State $D_i \gets \text{maxpool}(M_i, k=3, s=1, p=1)$ \Comment{Dilate mask}
        \State $D_j \gets \text{maxpool}(M_j, k=3, s=1, p=1)$ \Comment{Dilate mask}
        \If{$\sum_{x,y} (D_i \odot D_j)(x,y) > 0$}
            \State $\mathcal{E} \gets \mathcal{E} \cup \{(i, j)\}$ \Comment{Add edge if masks are adjacent}
            \State $e_{ij} \gets \text{MLP}([h_i \parallel h_j])$ \Comment{Compute edge feature}
        \EndIf
    \EndFor
\EndFor
\State \Return $\mathcal{G} = (\mathcal{V}, \mathcal{E})$, $H = [h_1, h_2, \ldots, h_K]^T$, $E = \{e_{ij} | (i, j) \in \mathcal{E}\}$
\end{algorithmic}
\end{algorithm}

For each class $c$, a soft binary mask $M_c = \mathbb{I}[\text{maxpool}(p_c) > \tau]$ is generated followed by thresholding. The feature tensor $\mathcal{D}_2$ is element-wise multiplied with $M_c$ to yield $F_c = \mathcal{D}_2 \odot M_c$. Each masked region is passed through a convolutional block and globally pooled via masked averaging to extract the tissue-specific node embedding $h_c \in \mathbb{R}^d$:
\begin{equation}
    h_c = \frac{\sum_{i,j} F_c(i,j)}{\sum_{i,j} M_c(i,j) + \varepsilon},
\end{equation}
where $\varepsilon$ is a small constant to prevent division by zero. All $K$ nodes form the node feature matrix $H \in \mathbb{R}^{K \times d}$. For each valid pair of tissue classes $(i, j)$ with spatially adjacent regions, we compute an edge embedding using a two-layer MLP on concatenated node features: $e_{ij} = \text{MLP}([h_i \Vert h_j])$. The resulting tissue graph is processed using a $L$-layer GNN, where at each layer $\ell$, node $i$ is updated as
\begin{equation}
    h_i^{(\ell+1)} = \sigma\left( \sum_{j \in \mathcal{N}(i)} W^{(\ell)} h_j^{(\ell)} \odot e_{ji} + b^{(\ell)} \right),
\end{equation}
with ReLU nonlinearity $\sigma$ and learnable weights $W^{(\ell)}$, $b^{(\ell)}$. To account for tissue classes absent in a given image, we replace the corresponding $h_i$ with a learned global embedding if $\sum M_i = 0$. The refined node embeddings $\{h_c^{(L)}\}$ are projected back to their corresponding spatial masks by broadcasting over $M_c$ and summing over all $K$ classes to construct the enhanced tensor $S \in \mathbb{R}^{d \times H_2 \times W_2}$:
\begin{equation}
    S = \sum_{c=1}^K h_c^{(L)} \otimes M_c.
\end{equation}
A $1\times1$ convolution with batch normalization projects $S$ to match the channel dimension of $\mathcal{D}_2$. The enriched spatial tensor $S$ is fused with $\mathcal{D}_2$ via residual addition and passed through the remaining decoder stages $\mathcal{D}_3$, $\mathcal{D}_4$, and $\mathcal{D}_5$, producing final segmentation logits $\hat{y}_{\text{final}}$. We train the model using a composite loss that combines predictions from both stages:
\begin{equation}
\mathcal{L}_{\text{total}} = \mathcal{L}_{\text{ce}}(\hat{y}_{\text{final}}, y) + \lambda \mathcal{L}_{\text{ce}}(\hat{y}_{\text{init}}, y),
\end{equation}
where $\lambda$ balances the auxiliary loss from the initial segmentation, and $y$ denotes the ground truth segmentation map containing pixel-wise class labels. The cross-entropy loss $\mathcal{L}_{\text{ce}}$ uses dynamic class weights computed per batch to counteract label imbalance across tissue types. Algorithm \ref{alg:graph_construction} outlines the procedure for constructing the tissue graph from initial segmentation probabilities and intermediate features.

\section{Experiments and Results}
\label{sec:experiments}

\subsection{Implementation Details}
\paragraph{Training Setup:} We adopt Adam optimizer with an initial learning rate of $1 \times 10^{-4}$, batch size of 4, and train for 150 epochs with early stopping. Data augmentation includes horizontal/vertical flipping and random rotations, applied as suggested by \cite{thomasBase}. Weighted cross-entropy loss was used to mitigate class imbalance, with auxiliary loss weighting $\lambda = 0.4$.

\paragraph{Dataset and Preprocessing:} We use the benchmark "Histopathology Non-Melanoma Skin Cancer Segmentation Dataset" from \cite{dataset}, which contains 290 whole slide images of non-melanoma skin cancers, comprising 140 BCC, 60 SCC, and 90 IEC cases. All slides are pre-processed using 10x, 5x, and 2x magnifications, with patches of size $256\times256$ extracted using the overlapping tiling strategy recommended in~\cite{thomasBase}. Following prior work as well as our obtained results, we adopt the 10x setting as our primary resolution due to its optimal balance between performance and efficiency.

\paragraph{Baselines:} We compare our method against four widely used segmentation architectures: Attention U-Net~\cite{attunet}, VGG U-Net~\cite{vgg,ronneberger2015u}, DeepLabV3+~\cite{deeplabv3}, and ResNet U-Net~\cite{thomasBase}. In addition, we include the Mix transformer encoders (MiT)-based Transformer framework from Imran et al.~\cite{imran2024transformer}, using the results as reported. All other models were re-trained on the dataset with identical data splits and training schedules for fairness.

\paragraph{Evaluation Metrics:} Accuracy, mean Intersection-over-Union (IoU), and Dice score are reported, with qualitative comparisons on representative samples containing as many classes as possible within each sample.

\begin{figure*}[t]
  \centering
  \includegraphics[width=0.95\textwidth]{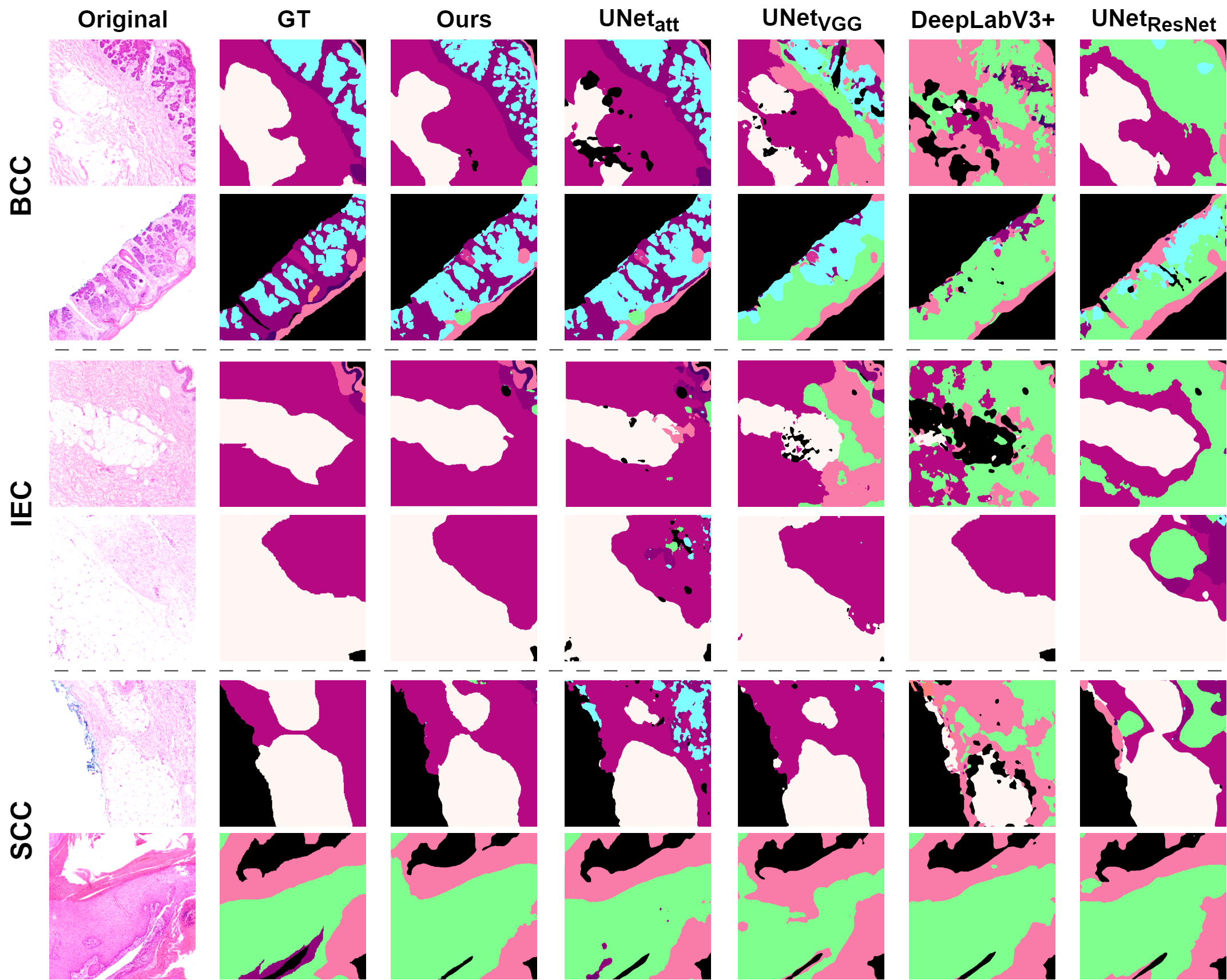}
  \caption{Qualitative comparison of segmentation results across non-melanoma skin cancer types: BCC, SCC, and IEC. Our method demonstrates improved localization of class boundaries and reduction in false positives (e.g., SCC) compared to others.}

  \label{fig:qual_comp}
\end{figure*}

\subsection{Qualitative and Quantitative Evaluations}

Fig.~\ref{fig:qual_comp} displays qualitative results across representative 
patches of BCC, SCC, and IEC. Looking at the BCC rows, baseline methods such 
as UNet$_{att}$~\cite{attunet} and DeepLabV3+~\cite{deeplabv3} frequently 
misclassify basal compartments as SCC (green) or background (black), 
fragmenting regions that should be spatially coherent. UNet$_{ResNet}$~\cite{thomasBase} 
handles gross structure reasonably well but loses precision near the 
epithelial-basal interface, producing ragged boundaries where our method 
maintains clean delineation. For IEC, the contrast is most visible in how 
competing methods struggle to separate the large epithelial mass from 
surrounding reticular dermis, often bleeding into adjacent classes, whereas 
our predictions hold a consistent boundary throughout. In SCC, where 
inflammation (INF), reticular dermis (RET), and follicular (FOL) structures 
sit in close proximity, baseline methods produce scattered misclassifications 
across the region, while our model preserves the spatial ordering of these 
tissue layers more faithfully.

Tab.~\ref{tab:quant_results} shows that our model achieves the highest mean 
IoU (0.7288) and Dice score (0.8163) across all evaluated methods. Although 
MiT~\cite{imran2024transformer} attains the highest accuracy (0.8310), its 
considerably lower mean IoU (0.653) reveals a bias toward dominant classes --- 
accuracy is heavily influenced by background pixels, which are abundant, while 
mean IoU equally penalizes poor segmentation of minority tissue regions 
regardless of their frequency. This distinction matters in histopathology, 
where clinically significant structures like BCC nests or keratin deposits 
occupy a small spatial footprint but carry diagnostic weight. Our model 
maintains a better balance between these two objectives, performing well on 
both majority and minority classes. UNet$_{ResNet}$~\cite{thomasBase} is the 
closest competitor but trails in both IoU and Dice, consistent with the 
boundary inconsistencies visible in Fig.~\ref{fig:qual_comp}. DeepLabV3+~\cite{deeplabv3} 
underperforms considerably across all metrics, which is unsurprising given 
that its atrous spatial pyramid pooling is designed for natural image 
semantics, where object boundaries are coarser and classes are more spatially 
separated. Histopathological tissue structures are fragmented and densely 
packed, and fixed dilation-based context aggregation is a poor fit for this 
setting compared to the spatially adaptive relational reasoning our model 
performs.

\begin{table}[t]
\centering
\scriptsize
\caption{\textbf{Quantitative comparison of methods at 10x magnification.} Metrics are averaged across tissue classes. MiT results are reported from~\cite{imran2024transformer}.}
\resizebox{\columnwidth}{!}{%
\begin{tabular}{lccc}
\toprule
Method & Accuracy $\uparrow$ & Mean IoU $\uparrow$ & Dice $\uparrow$ \\
\midrule
DeepLabV3+~\cite{deeplabv3} & 0.5061 & 0.4191 & 0.5038 \\
UNet$_{VGG}$~\cite{vgg,ronneberger2015u} & 0.6708 & 0.6002 & 0.7051 \\
UNet$_{att}$~\cite{attunet} & 0.7326 & 0.6326 & 0.7438 \\
UNet$_{ResNet}$~\cite{thomasBase} & 0.7368 & 0.6763 & 0.7674 \\
MiT~\cite{imran2024transformer} & \textbf{0.8310} & 0.6530 & - \\
\textbf{Ours} & 0.8106 & \textbf{0.7288} & \textbf{0.8163} \\
\bottomrule
\end{tabular}%
}
\label{tab:quant_results}
\end{table}

\begin{table}[t]
\centering
\scriptsize
\caption{\textbf{Comparison of our method across resolutions.} Lower magnification incurs moderate drop, supporting the choice of 10x for efficient yet accurate segmentation.}
\resizebox{\columnwidth}{!}{%
\begin{tabular}{lccc}
\toprule
Resolution & Accuracy $\uparrow$ & Mean IoU $\uparrow$ & Dice $\uparrow$ \\
\midrule
Ours (2x) & 0.7620 & 0.6691 & 0.7624 \\
Ours (5x) & 0.7920 & 0.6991 & 0.7924 \\
\textbf{Ours (10x)} & \textbf{0.8106} & \textbf{0.7288} & \textbf{0.8163} \\
\bottomrule
\end{tabular}
}
\label{tab:resolution_results}
\end{table}

\subsection{Computational Complexity}

The computational complexity of the NTRM can be analyzed in terms of both time and space requirements.

\paragraph{Time Complexity}

Let $N = H_2 \times W_2$ be the number of pixels in the feature map, $K$ be the number of tissue classes, and $d$ be the node embedding dimension. The time complexity of each component is:

\begin{itemize}
\item \textit{Region Proposal}: $O(K \cdot N)$ for generating masks for all classes.
\item \textit{Node Feature Extraction}: $O(K \cdot N \cdot d)$ for masking and pooling features.
\item \textit{Edge Formation}: $O(K^2 \cdot N)$ for checking adjacency between all pairs of classes, and $O(K^2 \cdot d^2)$ for computing edge features.
\item \textit{Graph Neural Network}: $O(L \cdot |E| \cdot d^2)$ for $L$ GNN layers with $|E|$ edges.
\item \textit{Feature Projection}: $O(K \cdot N \cdot d)$ for projecting node features to the spatial domain.
\end{itemize}

The total time complexity is $O(K \cdot N \cdot d + K^2 \cdot d^2 + L \cdot |E| \cdot d^2)$. Since $|E| \leq K^2$ and typically $K \ll N$, the time complexity is dominated by the $O(K \cdot N \cdot d)$ term for large images.

\paragraph{Space Complexity}

The space complexity is determined by the storage requirements for:

\begin{itemize}
\item \textit{Feature Maps}: $O(N \cdot d)$ for storing the projected features.
\item \textit{Masks}: $O(K \cdot N)$ for storing binary masks for all classes.
\item \textit{Graph}: $O(K \cdot d + |E| \cdot d)$ for storing node and edge features.
\end{itemize}

The total space complexity is $O(N \cdot d + K \cdot N + K \cdot d + |E| \cdot d)$, which simplifies to $O(N \cdot (d + K))$ for large images.

\paragraph{Comparison with Traditional Approaches}

Traditional convolutional approaches for segmentation have a time complexity of $O(N \cdot C^2 \cdot k^2)$, where $C$ is the number of channels and $k$ is the kernel size. Our TRM introduces an additional complexity of $O(K \cdot N \cdot d + K^2 \cdot d^2)$.

For histopathology images, $K$ is typically small (e.g., 12 in our dataset), making the overhead of the TRM reasonable compared to the base convolutional operations. Moreover, the TRM operates on a reduced spatial resolution (e.g., $1/8$ of the original image size), further reducing its computational impact.

\begin{figure}[t]
  \centering
  \includegraphics[width=\columnwidth]{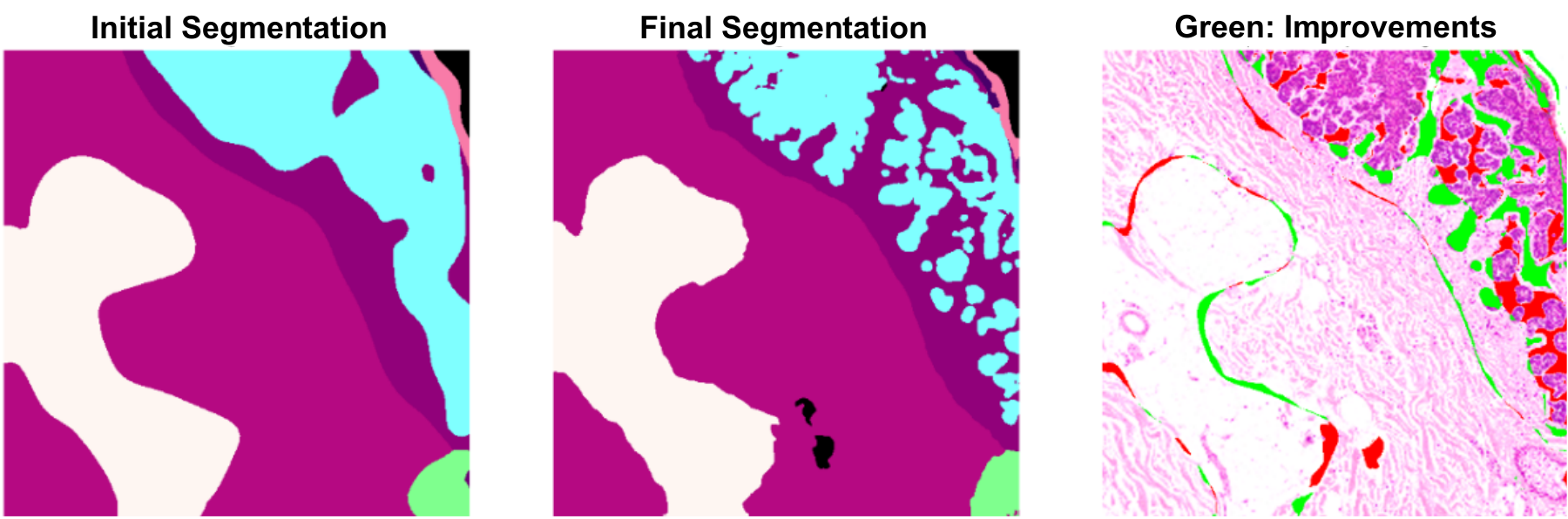}
  \caption{Impact of TRM: Left – initial segmentation from CNN; Center – final prediction after TRM; Right – improvement map overlay (green shows corrected predictions). TRM corrects major errors near BCC-reticular and RET-hypodermis interfaces.}
  \label{fig:trm_improvement}
\end{figure}

\subsection{Discussion}

\paragraph{Resolution Impact.} In Tab.~\ref{tab:resolution_results}, we evaluate our model at three magnification levels. The 10x setting consistently outperforms both 2x and 5x with only marginal computational cost, confirming prior findings~\cite{thomasBase} that 10x images contain sufficient granularity for tissue-level analysis while minimizing overhead. While 2x and 5x magnifications retain broader morphological context, they often lose fine-grained features critical for accurate tissue boundary delineation. The minor performance drop from 10× to 5× highlights the resilience of NTRM’s relational modeling under lower-resolution inputs, demonstrating robust generalization across acquisition settings while preserving high segmentation fidelity.

\paragraph{Effect of Tissue Relationship Modeling. }
We show the operational impact of the TRM module on segmentation refinement in Fig.~\ref{fig:trm_improvement}. The initial predictions, produced by the CNN decoder in isolation, show failure modes near complex boundaries - particularly at BCC-reticular interfaces and epithelial structures adjacent to keratin deposits. These misclassifications arise due to insufficient contextual reasoning across disjoint but functionally correlated tissue types. As depicted in the TRM pipeline (Fig.~\ref{fig:trm}), spatially contiguous regions are treated as graph nodes and connected via context-aware edges, allowing the network to explicitly reason over inter-tissue dependencies. The refined segmentation output captures granular class boundaries and suppresses spurious activations, as visually evident in the improvement overlay (right). It shows relational modeling helps in resolving biologically relevant structural ambiguities, rather than relying solely on visual proximity/texture.

Our results emphasize the importance of relational reasoning for complex tissue contexts. While conventional models rely primarily on texture and edge continuity, NTRM explicitly models biological adjacency and context through its graph-based TRM, which improves both boundary delineation and inter-class distinction. Our method outperforms all others on mean IoU and Dice, which are metrics more representative of segmentation fidelity in heterogeneous histological fields. Additionally, TRM’s modular design enables extension to other tissue-rich domains, offering interpretable and spatially structured refinement.

We do not include ablation studies in this work due to the tight architectural integration of our design. The TRM module functions as a relational refinement layer inherently coupled with the encoder-decoder structure, ensuring a coherent information flow. We refrain from fragmenting its components (spatially-aware masked pooling, graph-based reasoning, and projection), as this may not yield interpretable insights and would compromise structural coherence. Rather than isolating individual effects, we focus on demonstrating the collective efficacy of modeling inter-tissue dependencies holistically within a unified framework.

\section{Conclusion}
\label{sec:conclusion}

We proposed NTRM, a histopathology segmentation framework that models 
inter-tissue relationships through graph-based reasoning. By constructing 
a tissue-level graph over CNN-derived segmentations and refining predictions 
using relational embeddings, NTRM directly addresses a core limitation of 
texture-only models, which is their inability to encode the biological context that 
governs tissue organization in histological images. Rather than treating tissue 
regions as independent entities defined solely by visual appearance, NTRM 
explicitly learns how tissue types relate to one another spatially and 
functionally, which more closely mirrors the structured reasoning pathologists 
employ during diagnosis. Across non-melanoma skin cancer segmentation experiments spanning BCC, SCC, and IEC subtypes, NTRM consistently enhanced boundary delineation and class discrimination, achieving a mean IoU of 0.7288 and a Dice score of 0.8163 and outperforming all evaluated baselines by 4.9\% to 31.25\% in Dice. Qualitative comparisons show that relational modeling suppresses spurious activations at morphologically ambiguous boundaries, particularly where tissue types differ more in biological context than in visual texture. Taken together, these findings suggest that encoding structured biological knowledge alongside visual features may be as important as architectural advances in pushing segmentation performance.

\section*{Acknowledgements}
This work was supported by KAKENHI grant number JP24K15011.

{
    \small
    \bibliographystyle{ieeenat_fullname}
    \bibliography{main}
}

\clearpage
\setcounter{page}{1}
\maketitlesupplementary

\appendix
\renewcommand{\thesection}{A\arabic{section}}
\renewcommand{\thesubsection}{A\arabic{section}.\arabic{subsection}}
\renewcommand{\thefigure}{A\arabic{figure}}
\renewcommand{\thetable}{A\arabic{table}}

\setcounter{section}{0}
\setcounter{subsection}{0}
\setcounter{figure}{0}
\setcounter{table}{0}

\section{Extended Methodology}

This supplementary material provides a detailed mathematical formulation of the Neural Tissue Relation Modeling (NTRM) approach presented in the main paper. We extend the methodology by providing in-depth explanations of the tissue relation module, graph construction, message passing, and spatial projection mechanisms.

\subsection{Architectural Details}

The NTRM architecture extends a standard encoder-decoder framework with a novel Tissue Relation Module (TRM) that operates on intermediate features and initial segmentation predictions. While the main paper describes the overall structure, here we provide a more detailed mathematical description of each component.

The encoder extracts hierarchical features $\{\mathcal{E}_1, \mathcal{E}_2, \dots, \mathcal{E}_5\}$ from an input image $x \in \mathbb{R}^{3\times H \times W}$, where $\mathcal{E}_i \in \mathbb{R}^{C_i \times H_i \times W_i}$ represent features at different resolutions. The decoder produces intermediate features $\mathcal{D}_1 \in \mathbb{R}^{256 \times H_1 \times W_1}$ and $\mathcal{D}_2 \in \mathbb{R}^{128 \times H_2 \times W_2}$, which serve as inputs to the TRM.

The initial segmentation head is defined as a function $\phi: \mathbb{R}^{512 \times H_5 \times W_5} \rightarrow \mathbb{R}^{K \times H_0 \times W_0}$ that maps the bottleneck features $\mathcal{E}_5$ to a $K$-channel output, where $K$ is the number of tissue classes. The output is upsampled via bilinear interpolation $\mathcal{B}: \mathbb{R}^{K \times H_0 \times W_0} \rightarrow \mathbb{R}^{K \times H_2 \times W_2}$ to match the resolution of $\mathcal{D}_2$ and passed through a softmax function $\sigma$ to produce class probabilities:

\begin{equation}
p = \sigma(\mathcal{B}(\phi(\mathcal{E}_5))) \in \mathbb{R}^{K \times H_2 \times W_2}
\end{equation}

These probabilities, along with features $\mathcal{D}_2$, serve as inputs to the TRM, which we describe in detail in the following sections.

\subsection{Tissue Relation Module (TRM): Mathematical Formulation}

The Tissue Relation Module consists of four main components: region proposal, node feature extraction, edge formation, and graph neural network processing. Each component plays a critical role in modeling tissue relationships.

\subsubsection{Region Proposal}

The region proposal network takes initial segmentation probabilities $p \in \mathbb{R}^{K \times H_2 \times W_2}$ and extracts binary masks for each tissue class. For each class $c \in \{1, 2, \ldots, K\}$, we define a binary mask $M_c \in \mathbb{R}^{H_2 \times W_2}$ by thresholding the probability map:

\begin{equation}
M_c = \mathbb{I}[p_c > \tau]
\end{equation}

where $\mathbb{I}[\cdot]$ is the indicator function and $\tau$ is a threshold parameter (set to $0.5$ in our implementation). To enhance mask connectivity and address small gaps, we apply morphological operations approximated by max-pooling followed by thresholding:

\begin{equation}
M_c = \mathbb{I}[\text{maxpool}(p_c, k=3, s=1, p=1) > \tau]
\end{equation}

where $k$, $s$, and $p$ represent the kernel size, stride, and padding of the max-pooling operation, respectively.

\subsubsection{Node Feature Extraction}

Given the binary masks $\{M_1, M_2, \ldots, M_K\}$ and intermediate features $\mathcal{D}_2$, we extract node features for each tissue class. The features $\mathcal{D}_2$ are first processed through a convolutional layer $\psi: \mathbb{R}^{128 \times H_2 \times W_2} \rightarrow \mathbb{R}^{d \times H_2 \times W_2}$ to obtain refined features $F = \psi(\mathcal{D}_2)$, where $d$ is the node embedding dimension.

For each class $c$, we compute a masked representation $F_c = F \odot M_c$, where $\odot$ denotes the Hadamard product broadcast along the channel dimension. The node embedding $h_c \in \mathbb{R}^d$ is obtained via masked global average pooling:

\begin{equation}
h_c = \frac{\sum_{i,j} F_c(i,j)}{\sum_{i,j} M_c(i,j) + \varepsilon}
\end{equation}

where $(i,j)$ indexes spatial locations and $\varepsilon$ is a small constant ($10^{-6}$ in our implementation) to prevent division by zero when a class is absent. The complete node feature matrix is $H = [h_1, h_2, \ldots, h_K]^T \in \mathbb{R}^{K \times d}$.

\subsubsection{Edge Formation}

The edge formation process constructs a graph $\mathcal{G} = (\mathcal{V}, \mathcal{E})$ where vertices $\mathcal{V} = \{1, 2, \ldots, K\}$ represent tissue classes and edges $\mathcal{E} \subset \mathcal{V} \times \mathcal{V}$ represent spatial relationships.

We define two tissue classes $i$ and $j$ as spatially adjacent if their dilated masks have a non-zero intersection:

\begin{equation}
(i, j) \in \mathcal{E} \iff \sum_{x,y} (\text{dilate}(M_i) \odot \text{dilate}(M_j))(x,y) > 0
\end{equation}

where dilate is a morphological dilation operation with a $3 \times 3$ kernel. In practice, this is approximated using a max-pooling operation:

\begin{equation}
\text{dilate}(M) \approx \text{maxpool}(M, k=3, s=1, p=1)
\end{equation}

For each edge $(i, j) \in \mathcal{E}$, we compute an edge feature $e_{ij} \in \mathbb{R}^d$ using a two-layer MLP that operates on the concatenated node features:

\begin{equation}
e_{ij} = \text{MLP}([h_i \parallel h_j])
\end{equation}

where $\parallel$ denotes concatenation. The MLP consists of two linear layers with a ReLU activation in between:

\begin{equation}
\text{MLP}(x) = W_2(\text{ReLU}(W_1 x + b_1)) + b_2
\end{equation}

where $W_1 \in \mathbb{R}^{d \times 2d}$, $W_2 \in \mathbb{R}^{d \times d}$, $b_1 \in \mathbb{R}^{d}$, and $b_2 \in \mathbb{R}^{d}$ are learnable parameters.

\subsubsection{Graph Neural Network}

The graph neural network (GNN) processes the tissue graph to refine node features. We employ an $L$-layer GNN with residual connections and layer normalization. The update rule for node $i$ at layer $\ell$ is:

\begin{equation}
\tilde{h}_i^{(\ell)} = \text{LN}\left(h_i^{(\ell-1)} + \sum_{j \in \mathcal{N}(i)} \alpha_{ij} \cdot (W^{(\ell)} h_j^{(\ell-1)} \odot e_{ji})\right)
\end{equation}

\begin{equation}
h_i^{(\ell)} = \text{LN}\left(\tilde{h}_i^{(\ell)} + \text{FFN}(\tilde{h}_i^{(\ell)})\right)
\end{equation}

where $\mathcal{N}(i)$ denotes the neighborhood of node $i$ in the graph, $\alpha_{ij}$ is an attention coefficient, $W^{(\ell)} \in \mathbb{R}^{d \times d}$ is a learnable weight matrix, LN denotes layer normalization, and FFN is a feed-forward network consisting of two linear transformations with a ReLU activation in between:

\begin{equation}
\text{FFN}(x) = W_{\text{out}}(\text{ReLU}(W_{\text{in}} x + b_{\text{in}})) + b_{\text{out}}
\end{equation}

The attention coefficient $\alpha_{ij}$ is computed as:

\begin{equation}
\alpha_{ij} = \frac{\exp(a_{ij})}{\sum_{k \in \mathcal{N}(i)} \exp(a_{ik})}
\end{equation}

where $a_{ij} = \text{LeakyReLU}(q^T [W h_i^{(\ell-1)} \parallel W h_j^{(\ell-1)}])$ with $q \in \mathbb{R}^{2d}$ being a learnable attention vector. After $L$ GNN layers, we obtain refined node embeddings $H^{(L)} = [h_1^{(L)}, h_2^{(L)}, \ldots, h_K^{(L)}]^T \in \mathbb{R}^{K \times d}$.

\subsection{Global Tissue Embeddings}

In histopathology image analysis, not all tissue types are present in every image. To handle cases where certain tissue classes are absent, we introduce global tissue embeddings. For each class $c$ that is absent in an image (i.e., $\sum_{i,j} M_c(i,j) = 0$), we replace its node embedding $h_c$ with a learned global embedding $g_c \in \mathbb{R}^d$:

\begin{equation}
h_c = 
\begin{cases} 
h_c & \text{if } \sum_{i,j} M_c(i,j) > 0 \\
g_c & \text{otherwise}
\end{cases}
\end{equation}

The global embeddings $\{g_1, g_2, \ldots, g_K\}$ are learnable parameters that capture prior knowledge about tissue types and their relationships. They are initialized from a normal distribution $\mathcal{N}(0, 0.02)$ and updated during training.

The presence of global embeddings allows the model to reason about potential tissue interactions even when certain tissues are not present in the current image. This is particularly important for rare tissue types or when analyzing small image patches where not all tissues can be observed simultaneously. Mathematically, the global embeddings modify the node feature matrix $H$ by replacing absent tissue embeddings:

\begin{equation}
H' = H \odot P + G \odot (1 - P)
\end{equation}

where $P \in \{0, 1\}^{K \times d}$ is a binary presence matrix with $P_c = \mathbf{1}$ if tissue $c$ is present and $P_c = \mathbf{0}$ otherwise, and $G = [g_1, g_2, \ldots, g_K]^T \in \mathbb{R}^{K \times d}$ is the matrix of global embeddings.

\subsection{Feature Projection and Fusion}

After obtaining refined node embeddings $H^{(L)}$, we project them back to the spatial domain to produce enhanced features. For each tissue class $c$, we broadcast its embedding $h_c^{(L)}$ to the corresponding mask region:

\begin{equation}
S_c = h_c^{(L)} \otimes M_c \in \mathbb{R}^{d \times H_2 \times W_2}
\end{equation}

where $\otimes$ denotes the outer product that broadcasts the embedding to all spatial locations where $M_c = 1$. The enhanced spatial tensor $S$ is obtained by summing over all classes:

\begin{equation}
S = \sum_{c=1}^K S_c
\end{equation}

A $1 \times 1$ convolution with batch normalization is applied to project $S$ to match the channel dimension of $\mathcal{D}_2$:

\begin{equation}
S' = \text{BN}(\text{Conv}_{1 \times 1}(S))
\end{equation}

The enhanced features are fused with the original features via residual addition:

\begin{equation}
\mathcal{D}'_2 = \mathcal{D}_2 + S'
\end{equation}

The fused features $\mathcal{D}'_2$ are then passed through the remaining decoder stages to produce the final segmentation.

\subsection{Loss Function Analysis}

We train the model using a composite loss function that combines predictions from both the initial and final segmentation stages:

\begin{equation}
\mathcal{L}_{\text{total}} = \mathcal{L}_{\text{ce}}(\hat{y}_{\text{final}}, y) + \lambda \mathcal{L}_{\text{ce}}(\hat{y}_{\text{init}}, y)
\end{equation}

where $\lambda$ (set to $0.4$ in our implementation) balances the auxiliary loss from the initial segmentation, and $y$ denotes the ground truth segmentation map. The cross-entropy loss $\mathcal{L}_{\text{ce}}$ is weighted to account for class imbalance:

\begin{equation}
\mathcal{L}_{\text{ce}}(\hat{y}, y) = -\frac{1}{N} \sum_{n=1}^{N} \sum_{c=1}^{K} w_c \cdot y_{n,c} \log(\hat{y}_{n,c})
\end{equation}

where $N$ is the number of pixels, $w_c$ is the weight for class $c$, and $y_{n,c}$ and $\hat{y}_{n,c}$ are the ground truth and predicted probabilities for pixel $n$ and class $c$, respectively. The class weights $w_c$ are computed dynamically for each batch based on the frequency of each class:

\begin{equation}
w_c = \frac{N}{\sum_{n=1}^{N} \mathbb{I}[y_n = c] \cdot K}
\end{equation}

This formulation assigns higher weights to less frequent classes, helping the model learn from imbalanced data. The dual-stage loss serves multiple purposes:
\begin{enumerate}
\item It provides direct supervision to the initial segmentation head, ensuring meaningful features for the TRM.
\item It creates an auxiliary gradient path that facilitates training of the deeper layers.
\item It regularizes the model by encouraging consistent predictions at different stages.
\end{enumerate}

\section{Algorithmic Analysis}

\subsection{Edge Weight Computation}

The edge weights in the tissue graph represent the strength of the relationship between different tissue types. We compute these weights based on both spatial adjacency and feature similarity.

For each pair of adjacent tissue classes $(i, j) \in \mathcal{E}$, we define the edge weight $w_{ij}$ as:

\begin{equation}
w_{ij} = \sigma(e_{ij}^T W e_{ij})
\end{equation}

where $e_{ij}$ is the edge feature, $W \in \mathbb{R}^{d \times d}$ is a learnable weight matrix, and $\sigma$ is the sigmoid function that maps the weight to the range $(0, 1)$. To capture the asymmetric nature of tissue relationships (e.g., tumor cells might influence surrounding tissues differently than vice versa), we allow $w_{ij} \neq w_{ji}$ by computing them separately.

\subsection{Handling Boundary Cases}

Histopathology images often contain tissue boundaries where multiple tissue types meet. These regions require special attention in the graph construction process. We employ two strategies to handle boundary cases:

1. \textit{Soft Mask Assignment}: Instead of using hard binary masks, we can use soft masks based on the probability values $p_c$. This allows a pixel to contribute to multiple tissue nodes proportionally to its class probabilities.

2. \textit{Boundary-Aware Edge Features}: For edges that connect tissues with a significant boundary, we compute additional boundary-specific features:

\begin{equation}
b_{ij} = \frac{\sum_{x,y} (M_i \odot \text{dilate}(M_j))(x,y)}{\sum_{x,y} M_i(x,y)}
\end{equation}

This boundary ratio $b_{ij}$ represents the fraction of tissue $i$ that is adjacent to tissue $j$. It is incorporated into the edge feature computation:

\begin{equation}
e_{ij} = \text{MLP}([h_i \parallel h_j \parallel b_{ij} \parallel b_{ji}])
\end{equation}

By explicitly modeling boundary information, the graph can better capture tissue interactions at interface regions, which are often clinically significant (e.g., tumor invasions).

\subsection{Forward Pass Algorithm}

Algorithm \ref{alg:forward_pass} outlines the complete forward pass of the NTRM model.

\begin{algorithm}
\caption{NTRM Forward Pass}
\label{alg:forward_pass}
\begin{algorithmic}[1]
\Require Input image $x \in \mathbb{R}^{3 \times H \times W}$
\Ensure Final segmentation $\hat{y}_{\text{final}} \in \mathbb{R}^{K \times H \times W}$

\State $\{\mathcal{E}_1, \mathcal{E}_2, \ldots, \mathcal{E}_5\} \gets \text{Encoder}(x)$ \Comment{Extract encoder features}
\State $\mathcal{D}_1 \gets \text{Decoder1}(\mathcal{E}_5, \mathcal{E}_4)$ \Comment{First decoder block}
\State $\mathcal{D}_2 \gets \text{Decoder2}(\mathcal{D}_1, \mathcal{E}_3)$ \Comment{Second decoder block}

\State $\hat{y}_{\text{init}} \gets \phi(\mathcal{E}_5)$ \Comment{Initial segmentation}
\State $p \gets \sigma(\mathcal{B}(\hat{y}_{\text{init}}))$ \Comment{Upsample and apply softmax}

\State $\mathcal{G}, H, E \gets \text{ConstructGraph}(p, \mathcal{D}_2)$ \Comment{Construct tissue graph}

\For{$\ell = 1$ to $L$}
    \For{$i = 1$ to $K$}
        \State $\tilde{h}_i^{(\ell)} \gets h_i^{(\ell-1)} + \sum_{j \in \mathcal{N}(i)} \alpha_{ij} \cdot (W^{(\ell)} h_j^{(\ell-1)} \odot e_{ji})$
        \State $\tilde{h}_i^{(\ell)} \gets \text{LayerNorm}(\tilde{h}_i^{(\ell)})$
        \State $h_i^{(\ell)} \gets \tilde{h}_i^{(\ell)} + \text{FFN}(\tilde{h}_i^{(\ell)})$
        \State $h_i^{(\ell)} \gets \text{LayerNorm}(h_i^{(\ell)})$
    \EndFor
\EndFor

\State $S \gets \sum_{c=1}^K h_c^{(L)} \otimes M_c$ \Comment{Project to spatial domain}
\State $S' \gets \text{BN}(\text{Conv}_{1 \times 1}(S))$ \Comment{Project to match channels}
\State $\mathcal{D}'_2 \gets \mathcal{D}_2 + S'$ \Comment{Fuse features}

\State $\mathcal{D}_3 \gets \text{Decoder3}(\mathcal{D}'_2, \mathcal{E}_2)$ \Comment{Third decoder block}
\State $\mathcal{D}_4 \gets \text{Decoder4}(\mathcal{D}_3, \mathcal{E}_1)$ \Comment{Fourth decoder block}
\State $\mathcal{D}_5 \gets \text{Decoder5}(\mathcal{D}_4)$ \Comment{Fifth decoder block}
\State $\hat{y}_{\text{final}} \gets \text{FinalConv}(\mathcal{D}_5)$ \Comment{Final segmentation}

\State \Return $\hat{y}_{\text{final}}$, $\hat{y}_{\text{init}}$
\end{algorithmic}
\end{algorithm}

\section{Optimization Details}

\subsection{Learning Rate Scheduling}

We employ a learning rate scheduling strategy to improve convergence, which reduces the learning rate when the validation loss plateaus:

\begin{equation}
\text{lr}_{\text{new}} = 
\begin{cases} 
\text{lr}_{\text{old}} \cdot \gamma & \text{if no improvement for } \text{patience} \text{ epochs} \\
\text{lr}_{\text{old}} & \text{otherwise}
\end{cases}
\end{equation}

where $\gamma = 0.5$ is the reduction factor and patience = 5 epochs. This scheduling strategy allows the model to make large updates initially and then fine-tune as training progresses.

\subsection{Weight Initialization}

Proper weight initialization is crucial for training deep networks. We employ the following initialization schemes:

\begin{itemize}
\item \textit{Convolutional Layers}: Weights are initialized using Kaiming initialization with a normal distribution:

\begin{equation}
W \sim \mathcal{N}(0, \sqrt{\frac{2}{(1 + a^2) \cdot \text{fan\_in}}})
\end{equation}

where $a$ is the negative slope of the leaky ReLU (or $a = 0$ for standard ReLU) and fan\_in is the number of input units.

\item \textit{Graph Neural Network}: Edge weights and attention parameters are initialized from a uniform distribution:

\begin{equation}
W \sim \mathcal{U}(-\sqrt{\frac{6}{d_{\text{in}} + d_{\text{out}}}}, \sqrt{\frac{6}{d_{\text{in}} + d_{\text{out}}}})
\end{equation}

where $d_{\text{in}}$ and $d_{\text{out}}$ are the input and output dimensions.

\item \textit{Global Tissue Embeddings}: Initialized from a normal distribution with $\mu = 0$ and $\sigma = 0.02$:

\begin{equation}
g_c \sim \mathcal{N}(0, 0.02)
\end{equation}
\end{itemize}

\subsection{Normalization Configurations}

Batch normalization is applied after convolutional layers to stabilize training. For the TRM, we use layer normalization instead of batch normalization for the graph neural network, as it is more suitable for graph-structured data where the batch size may be variable. The batch normalization layers use the following configuration: momentum of 0.1, epsilon of \(1 \times 10^{-5}\), and affine parameters enabled. During inference, we use the running statistics accumulated during training for batch normalization.

\end{document}